\documentclass[conference]{IEEEtran}
\usepackage{amsmath,amssymb,amsfonts}
\usepackage{algorithmic}
\usepackage{bm}
\usepackage{textcomp}
\usepackage{xcolor}
\usepackage{booktabs}
\usepackage{multirow}
\usepackage{subcaption}
\usepackage{diagbox}
\usepackage{graphicx}
\usepackage{tikz}
\usepackage{cite}
\usepackage{hyperref}

\def\BibTeX{{\rm B\kern-.05em{\sc i\kern-.025em b}\kern-.08em
    T\kern-.1667em\lower.7ex\hbox{E}\kern-.125emX}}

\begin{document}

\title{
    Expert-Adaptive Medical Image Segmentation\\
}

\author{
    \IEEEauthorblockN{Binyan Hu}
    \IEEEauthorblockA{
        \textit{Department of Computing Technologies} \\
        \textit{Swinburne University of Technology}\\
        Hawthorn, Australia \\
        bhu@swin.edu.au
    }
    \and
    \IEEEauthorblockN{A. K. Qin}
    \IEEEauthorblockA{
        \textit{Department of Computing Technologies} \\
        \textit{Swinburne University of Technology}\\
        Hawthorn, Australia \\
        kqin@swin.edu.au
    }
}

\maketitle

\begin{abstract}
Medical image segmentation (MIS) plays an instrumental role in medical image analysis, where considerable effort has been devoted to automating the process. Currently, mainstream MIS approaches are based on deep neural networks (DNNs), which are typically trained on a dataset with annotations produced by certain medical experts. In the medical domain, the annotations generated by different experts can be inherently distinct due to complexity of medical images and variations in expertise and post-segmentation missions. Consequently, the DNN model trained on the data annotated by some experts may hardly adapt to a new expert. In this work, we evaluate a customised expert-adaptive method, characterised by multi-expert annotation, multi-task DNN-based model training, and lightweight model fine-tuning, to investigate model’s adaptivity to a new expert in the situation where the amount and mobility of training images are limited. Experiments conducted on brain MRI segmentation tasks with limited training data demonstrate its effectiveness and the impact of its key parameters.
\end{abstract}


\section{Introduction}

Medical image segmentation (MIS) plays a crucial role in the workflow of medical images-based diagnosis and theranostics, marking out the regions of interest in a medical image for follow-up medical actions. Nowadays, MIS based on machine learning (ML) has been extensively studied and widely applied to relieve medical specialists (experts) from the labour-intensive and error-prone manual segmentation process. Among existing ML-based MIS techniques, those based on deep neural networks (DNNs) \cite{ronneberger2015u} have become the mainstream technology, demonstrating remarkable performance advantages in various MIS tasks \cite{isensee2021nnu}.

\begin{figure*}[htbp]
    \centering
    \foreach \iexpert in {0,1,2,3,4,5,6} {
        \includegraphics[width=0.13\linewidth]{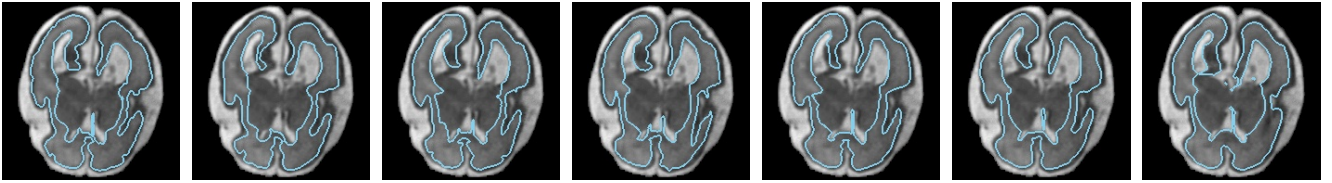}
        \hspace*{-0.2cm}
    }
    \caption{Illustration of the brain contours results created by seven different experts on the same MRI. It can be observed that the regions of interest (ROIs) marked out by different experts are largely consistent, though the contours themselves may vary greatly in different locations. Such contour variations may not impact disease diagnosis, which depends more on features within the ROIs, but may significantly influence the segmentation model trained by using them as annotations.}
    \label{fig:vis_brain_growth}
\end{figure*}

In DNN-based MIS techniques, a DNN model is typically trained on a collection of annotated medical images. The annotations, often in the form of segmentation masks or contours, are generated by certain medical experts. After the model is trained, it will be deployed for use by various medical experts who are often not among those annotating the training data. Different from generic image segmentation applications, annotations in MIS are highly expert-specific. On one hand, medical images contain complex features, e.g., blurry regions, transition zones, and partial volume effects, which pose a great challenge to creating annotations. On the other hand, medical experts may have varying expertise and skills for reading and annotating a medical image and may also target distinct post-segmentation tasks. As a result, the segmentation annotations produced by different medical experts on the same medical image can be inherently different but fully reasonable. \autoref{fig:vis_brain_growth} illustrates this phenomenon via the brain segmentation contours created by seven different experts on the same MRI image.

The expert-specific feature of the annotations in MIS may degrade the effectiveness of conventional DNN-based MIS techniques because the model trained on a collection of expert-specific annotated data is inherently biased towards those experts who have annotated the training data. Accordingly, the segmentation results generated by this model may not satisfy a new expert, as experimentally demonstrated in \autoref{sec:occur_expert_specific}. An intuitive solution is to re-train the model using medical images with annotations created by the new expert. However, obtaining sufficient annotated images from a new expert to effectively re-train the model to cater for this expert is seldom feasible in practice.

In this work, we investigate an existing expert-adaptive MIS method \cite{nichyporuk2022rethinking} which leverages multi-expert annotated data to train a multi-task DNN-based segmentation model that is capable of efficiently adapting to a new expert via lightweight model fine-tuning on limited annotated data from this expert. Specifically, the training data are first created by annotating different cohorts of medical images by different experts. Then, a multi-task DNN-based segmentation model is trained on such training data. When a new expert uses the trained model, the model, tailored for a single expert, will be fine-tuned on a small number of annotated medical images from this new expert. This lightweight fine-tuning step can efficiently adapt the model to cater for the new expert. In practice, the image data available for model training is often limited. Also, it is privacy-risky and costly to outsource the annotation of own data to other parties. Given these considerations, we customise this method by letting multiple different experts (possibly from the same party) to annotate every training image (aiming to better handle the limitation of training data availability and mobility) and applying the full model training in the fine-tuning stage (to better leverage the limited annotated data from the new expert), and evaluate the customised method on brain MRI segmentation tasks with limited training data. Experimental results demonstrate its effectiveness and the impact of its key parameters.

The rest of this paper is organised as follows. \autoref{related_work} describes the related work. The method is detailed in \autoref{sec:method}. \autoref{sec:experiments} presents experimental results. The conclusions with the future work are summarised in \autoref{conclusions}.

\section{Related Work}
\label{related_work}



The inherent variation in segmentation annotations from different experts poses a significant challenge to conventional image segmentation techniques, which typically assume the existence of an exclusive ground truth. To address this issue, many existing works based on multi-expert annotation have been proposed.

One category of methods is based on label fusion like \cite{warfield2004simultaneous}, which intends to combine different annotations from different experts with respect to one image into a single annotation to accommodate the annotation variation. Such methods may produce a consistent segmentation output that pursues the best-match with the annotation experts but can hardly cater for a new expert. Another category of methods generates the segmentation result in a probabilistic manner to accommodate the inherent expert-specific variation. For example, in \cite{yang2022uncertainty}, multi-expert annotation is used to define an uncertainty map to quantify the variation of segmentation annotations produced by different experts, where regions with high variation among experts are highlighted. An MIS model is then trained to predict such an uncertainty map for an input image in addition to the segmentation mask. Probabilistic U-Net \cite{kohl2018probabilistic} learns to output a distribution from which the segmentation result for an input image can be sampled in a certain way. Additionally, some works propose to produce expert-specific outputs \cite{ji2021learning, nichyporuk2022rethinking}. They leverage multi-expert annotation to train a multi-task learning model, which can output multiple segmentation results for one image, with each result corresponding to an expert who annotates the training data. 

In \cite{nichyporuk2022rethinking}, multiple cohorts of images are annotated by different experts, with one image annotated by only one expert. Then, a multi-task DNN-based segmentation model is trained on the thus created training set to produce the segmentation result for each expert, where each expert corresponds to one of multiple branches in the model with each branch differing in certain task-specific layers and sharing the remaining layers. Finally, the trained model is tailored to be used by a new expert, where the shared layers are inherited with their parameters frozen, and the task-specific layers are inherited from a selected task branch in the trained model and then fine-tuned by 10 annotated images from the new expert. In this work, we customise the multi-expert annotation and the model fine-tuning for the new expert components in this method, by using multiple different experts to annotate every training image (to better deal with the situation of limited training data availability and mobility) and applying the full model training in the fine-tuning stage (to make better use of the limited annotated data from the new expert).

\section{Method}
\label{sec:method}

\begin{figure*}[htbp]
    \centering
    \includegraphics[width=\textwidth]{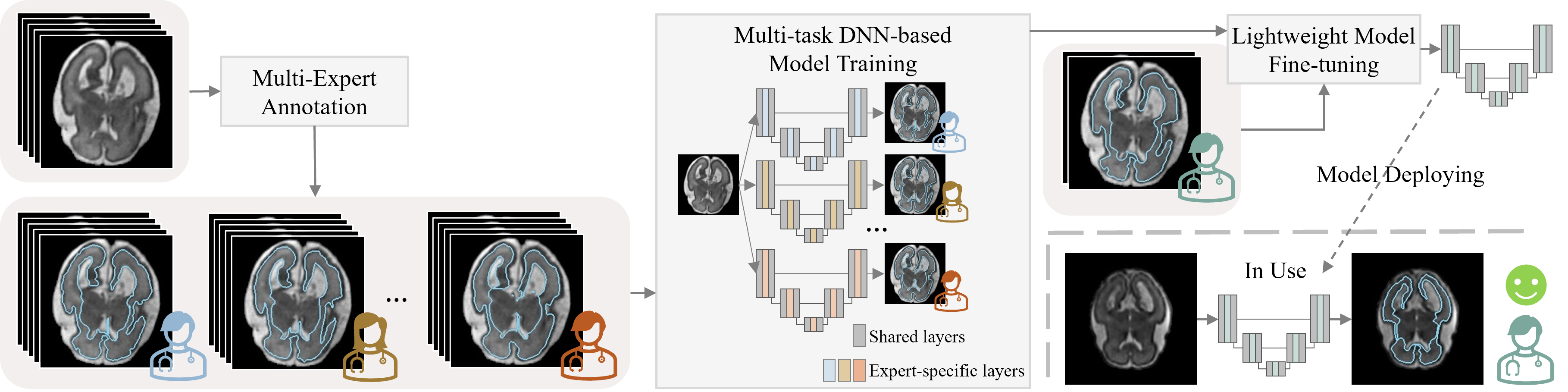}
    \caption{An illustration of the customised expert-adaptive medical image segmentation method, composed of multi-expert annotation, multi-task DNN-based model training, and lightweight model fine-tuning (based on a small number of annotated data from a new expert).    
    }
    \label{fig:multiexpert_framework}
\end{figure*}

The expert-adaptive MIS method, customised from \cite{nichyporuk2022rethinking}, is illustrated in \autoref{fig:multiexpert_framework}, which leverages multi-expert annotated images to train a segmentation model that is capable of efficiently adapting to a new expert via lightweight model fine-tuning on a small number of images associated with annotations from this expert. The following describes the working principle of this method.

Given a set of $N_{tr}$ medical images $\{\bm{x}_i\}_{i=1}^{N_{tr}}$ available for training, multi-expert annotation is first conducted by letting $N_{exp}$ different medical experts, denoted by $\{Exp_r\}_{r=1}^{N_{exp}}$ to annotate each image by producing segmentation contours. As a result, each training image is associated with $N_{exp}$ annotations, yielding the training set $\{\bm{x}_i,\{\bm{y}_i^r\}_{r=1}^{N_{exp}}\}_{i=1}^{N_{tr}}$.
Then, a multi-task DNN-based segmentation model is trained on this training set. Specifically, the conditioned instance normalisation (CIN) approach \cite{nichyporuk2022rethinking} is used to build a multi-task learning model based on the conventional U-Net architecture, which has demonstrated good performance in multi-task learning based MIS. The U-Net model is composed of repetitions of the sequential stack of convolution, instance normalisation, and activation layers. CIN makes the instance normalisation layers expert-specific, parameterised by 
$\{\bm{\theta}_r\}_{r=1}^{N_{exp}}$, with the remaining layers shared by all experts, parameterised by $\bm{\theta}_{sh}$. Accordingly, each expert corresponds to a specific branch, as shown in \autoref{fig:multiexpert_framework}.
During the training, the output with respect to each expert aims to approximate the segmentation annotation by the corresponding expert. The training loss function is defined as follows:
\begin{equation}
    l_{tr}=\sum_{i=1}^{N_{tr}}{\sum_{r=1}^{N_{exp}}{\mathcal{L}(\bm{y}_i^r,f(\bm{x}_i;\bm{\theta}_{sh},\bm{\theta}_r)})},
\end{equation}
where $f(\cdot;\bm{\theta}_{sh},\bm{\theta}_r)$ denotes the branch of the model for $Exp_r$, and $\mathcal{L}$ indicates the Dice loss function. The model parameters ${\bm{\theta}_{sh}}$ and $\{\bm{\theta}_r\}_{r=1}^{N_{exp}}$ are trained to minimise $l_{tr}$. After the training, the parameters of the shared part capture common features shared across all experts while those of the expert-specific parts carry unique features dedicated to producing expert-specific segmentation outputs. To enable the swift adaption of the trained model to a new expert, the shared part $\bm{\theta}_{sh}$ is kept while inheriting its parameters from the trained model, and combined with a re-initialised expert-specific part, parameterised by $\bm{\theta}_u$, to create a model which is fully fine-tuned on a small number of $N_{ft}$ annotated data from the new expert $\{\bm{x}_i,\bm{y}_i^u\}_{i=1}^{N_{ft}}$ to quickly achieve the desired performance that satisfies this expert. The loss function in the fine-tuning stage is formulated as:
\begin{equation}
    l_{ft}=\sum_{i=1}^{N_{ft}}{\mathcal{L}(\bm{y}_i^u,f(\bm{x}_i;\bm{\theta}_{sh},\bm{\theta}_u))},
\end{equation}
The model parameters, i.e., ${\bm{\theta}_{sh}}$ and $\bm{\theta}_u$, are trained to minimise $l_{ft}$. The major distinction bewteen the original method \cite{nichyporuk2022rethinking} and its customised version lies in the way of multi-expert annotation and the strategy of fine-tuning the model for a new expert.

\section{Experiments}
\label{sec:experiments}

We evaluate the performance of the customised expert-adaptive method on brain MRI segmentation tasks with limited training data, created from a publicly available brain MRI segmentation dataset with multi-expert annotations, and investigate the impact of its key parameters. The following experiments aim to demonstrate:
\begin{enumerate}
    \item The occurrence of the expert-specific issue when a new expert uses a model trained on the data annotated by other experts, demonstrating the necessity of expert-adaptive methods (\autoref{sec:occur_expert_specific}).
    \item The effectiveness of the evaluated expert-adaptive method (\autoref{sec:num_ann})
    \item The influence of the number of experts participating in multi-expert annotation on the performance of the evaluated expert-adaptive method (\autoref{sec:num_experts})
\end{enumerate}

\subsection{Dataset}
\label{sec:dataset}

We conduct our experiments based on the brain-growth dataset from the QUBIQ (Quantification of Uncertainties in Biomedical Image Quantification) challenge \cite{menze2020quantification}. The raw dataset contains 39 cases (34 for training and 5 for testing) for brain segmentation on MRI. Each image has 7 annotation masks of the brain, provided by 7 experts, $\{Exp_1, Exp_2, ..., Exp_7\}$, and a visual comparison of the annotations of one case is shown in \autoref{fig:vis_brain_growth}.

To simulate the expert adaptation scenario, we partition the experts into two groups: the experts who may annotate the multi-expert dataset for model training ($Exp_1$-$Exp_5$) and the new experts who will use the trained model ($Exp_6$ and $Exp_7$). The reason for selecting $Exp_6$ and $Exp_7$ as the new experts is that: $Exp_6$ represents an expert who produces similar annotations to certain annotation experts ($Exp_1$-$Exp_5$), whereas $Exp_7$ represents an expert who produces relatively different annotations from all annotation experts, as will be explained in detail in \autoref{sec:occur_expert_specific}. In practice, the condition of the new expert's annotations would be unknown, so our experimental setting aims to cover both cases.

As the following experiments may involve model training on multi-expert annotations and model fine-tuning on a small number of samples, we may use only a part of the 5 annotation experts and a part of the total 34 samples in the training set, which means the creation of sub-sets through the sampling annotation experts or training data. However, the performance can be strongly influenced by the specific combinations of the experts and the data samples, which may overshadow the impact of the factors we aim to investigate: the number of experts and the number of data samples. To tackle this issue, we adopt the following sub-set sampling method within the multi-run experimental setting.

First, to mitigate the impact of specific combinations of experts in the training stage, we experiment with every combination of experts, e.g., $C_5^4=5$ combinations for the experiment involving 4 experts.
Second, to mitigate the impact of specific combinations of training data in the fine-tuning stage, we adopt the following data sampling strategy to yield 10 ways of sampling a certain number of data from the whole training set, aiming to sample the data evenly in a deterministic manner. We first specify 10 starting indices evenly set based on the cardinality of the dataset, i.e., 1, 4, 7, ...28. Then, in each way of sampling, consecutive indices from the starting index are sampled. For example, when sampling 10 data, the 1$^{st}$ way yields samples 1-10, the 2$^{nd}$ way yields samples 4-13, ..., and the $10^{th}$ way yields samples 28-34 and 1-3. Additionally, in each way of data sampling, the data sampled with a larger number always contain those sampled with a smaller number.
Putting together, when involving 4-expert annotations for training and using 10 samples for fine-tuning, in total 50 runs of experiments are conducted, as a result of sampling both the experts and the training data.

After finishing all the experiments, the results from the same way of data sampling but different combinations of experts are averaged to produce a single result. Thus, the variance incurred by specific combinations of experts is effectively reduced. Finally, the experiment involving any number of experts and any number of data samples will uniformly yield 10 statistical results, each corresponding to a specific way of data sampling.

\subsection{Experimental Setup}
\label{sec:setup}

The model architecture is based on the U-Net \cite{ronneberger2015u} with the backbone of ResNet-18 \cite{he2016deep}, which is widely adopted for MIS tasks. For training with the expert-adaptive method, CIN \cite{nichyporuk2022rethinking} is used to construct a multi-task learning model based on the U-Net architecture, as detailed in \autoref{sec:method}. Otherwise, a conventional U-Net model is used. The training data are augmented via random translation, zooming, rotation, Gaussian noise, Gaussian blur, and brightness jittering following \cite{isensee2021nnu}. All images are cropped to a size of $192\times192$ for training and testing. The batch size is set to 16. We use the Rectified Adam optimiser \cite{Liu2020On} for model training. The learning rate is initialised as 0.001 and decayed throughout the training process by the polynomial annealing scheduler with a power rate of 0.9. Every model is trained for 5000 iteration steps in the training stage and if applicable, 1000 iteration steps in the fine-tuning stage.

We evaluate the segmentation performance with three metrics commonly used in the medical domain: 1) the Dice score (\textbf{Dice}), 2) average symmetric surface distance (\textbf{ASSD}), and 3) 95\% Hausdorff distance (\textbf{95HD}), and perform t-tests to assess the significance of difference. Specifically, in the experiments in the following section, we use unpaired t-tests. In the two experiments in the latter two sections, we perform paired t-tests because each statistical result in the multi-run corresponds to a certain way of data sampling due to our sub-set sampling method.

\subsection{Occurrence of the Expert-specific Issue}
\label{sec:occur_expert_specific}

We perform experiments to show the occurrence of the expert-specific issue when a new expert uses a model trained on the data annotated by other experts. In each experiment, we train a model with the training set annotated by one expert from $Exp_1-Exp_7$ and evaluate the model with the test set annotated by $Exp_6$ or $Exp_7$. Experiments are conducted on all the train-test expert pairs.

\begin{table}[htbp]
    \centering
    \caption{Comparision of segmentation results for $Exp_6$ and $Exp_7$, respectively, using the models trained on the data annotated by each of seven experts $Exp_1-Exp_7$. Metrics of  Dice, ASSD, and 95\% HD metrics are reported. Each cell reports the performance of a model trained on the training set annotated by one expert. Every experiment is run 10 times, and the mean metrics are reported. For the metrics in each column, the best performer and those not significantly different from the best (unpaired t-test with significance level of $\alpha=0.05$) are highlighted in \textbf{bold}.}
    \label{tab:occur_expert_specific}

    \begin{subtable}{\linewidth}
        \centering
        \caption{Tested on $Exp_6$.}
        \begin{tabular}{c|c|c|c}
            \toprule
            Train       & Dice (\%) $\uparrow$ & ASSD $\downarrow$ & 95HD $\downarrow$    \\
            \midrule
            $Exp_1$     &         86.61     &           1.84  &          5.60  \\
            $Exp_2$     &         86.66     &           1.93  &          6.57  \\
            $Exp_3$     &         87.31     &           1.76  &          6.29  \\
            $Exp_4$     &         86.92     &           1.76  &          6.05  \\
            $Exp_5$     &         88.60     &           1.64  &  \textbf{5.17} \\
            $Exp_6$     & \textbf{89.37}    &   \textbf{1.56} &  \textbf{4.87} \\
            $Exp_7$     &         83.29     &           2.15  &          6.88  \\
            \bottomrule
        \end{tabular}
    \end{subtable}
    \vspace{1em}\\
    \begin{subtable}{\linewidth}
        \centering
        \caption{Tested on $Exp_7$.}
        \begin{tabular}{c|c|c|c}
            \toprule
            Train       & Dice (\%) $\uparrow$ & ASSD $\downarrow$ & 95HD $\downarrow$    \\
            \midrule
            $Exp_1$     &         86.11     &          1.86  &          6.63  \\
            $Exp_2$     &         85.23     &          2.16  &          7.93  \\
            $Exp_3$     &         85.44     &          2.04  &          8.01  \\
            $Exp_4$     &         85.16     &          1.95  &          7.03  \\
            $Exp_5$     &         86.34     &          1.85  &          7.03  \\
            $Exp_6$     &         85.96     &          1.90  &          7.16  \\
            $Exp_7$     & \textbf{90.91}    &  \textbf{1.21} &  \textbf{3.91} \\
            \bottomrule
        \end{tabular}
    \end{subtable}
\end{table}

Results are shown in \autoref{tab:occur_expert_specific}. It can be observed that across all three metrics, for both $Exp_6$ and  $Exp_7$, the best performer is often the model trained on annotations of themselves, and other models perform significantly worse, with the only exception of the 95HD of the model trained on $Exp_5$ and tested on $Exp_6$. When using a model trained on another expert's annotations, the Dice, ASSD, and 95HD metrics degrade by 0.77\%-6.08\%, 0.08-0.59, and 0.3-2.01, respectively on $Exp_6$. On $Exp_7$, the three metrics degrade by 4.57\%-5.75\%, 0.64-0.95, and 2.72-4.1, respectively when using a model trained on other experts. These observations indicate that due to expert-specific annotations, an MIS model trained by a single expert's annotations often cannot readily cater for a new expert. Therefore, adapting the model to the new expert is necessary.
Notably, there are much larger performance drops observed in the results tested on $Exp_7$, compared with those of $Exp_6$. For example, on $Exp_6$, $Exp_5$'s model yields only a slight performance drop of 0.77\%, 0.08, and 0.3 on the three metrics, whereas on $Exp_7$, all the other experts' models incur large performance gaps. This suggests that the annotations produced by $Exp_7$ differ from all the other experts to a large extent. In the application, the new expert's annotations can potentially vary much from the annotations used for model training, in which case adapting the model to them tends to be harder.

\subsection{Effectiveness of the Expert-Adaptive Segmentation Method}
\label{sec:num_ann}

We investigate the effectiveness of the expert-adaptive method in enabling efficient adaptation to a new expert with a small number of annotations provided. To this end, experiments are conducted to compare the method with the baseline method which re-trains a model solely with the annotated samples provided by the new expert. In addition, we experiment with different numbers of annotations provided by the new expert, aiming to analyse under what conditions the expert-adaptive method is effective. Specifically, in the training stage, we use the dataset annotated by 3 experts to train the model. In the fine-tuning stage, the model is fine-tuned for $Exp_6$ or $Exp_7$ using 5, 10, 15, 20, 25, 30, and 34 annotated samples, respectively in different experiments. By contrast, the baseline method solely trains a randomly initialised model using the annotated samples from the new expert.

\begin{table}[htbp]
    \centering
    \caption{
        Comparison of segmentation performance on a new expert ($Exp_6$ or $Exp_7$) with different numbers of annotated samples. Each column under "w/" lists the results obtained by training the model with a dataset annotated by 3 experts from $Exp_1$-$Exp_5$ and then adapting it to the new expert with different numbers of annotated samples, and each column under "w/o" lists the results obtained by training a model from random initialisation solely with the new expert's annotated samples. Every experiment yields 10 statistical runs, and the mean metrics of Dice score, ASSD, and 95HD tested on the new expert's annotations are reported. The best performers in terms of mean value and those not significantly different from the best (paired t-test with significance level of $\alpha=0.05$), are highlighted in \textbf{bold}. For each metric in the "w/" column, if it is significantly better than its counterpart in the "w/o" column, it is \underline{underlined}.
    }
    \label{tab:n_ann}
    \begin{subtable}{\linewidth}
        \centering
        \caption{Adapt to $Exp_6$.}
        \begin{tabular}{c|cc|cc|cc}
            \toprule
            \multirow{2}{*}{\diagbox[width=3cm]{\# Samples}{Multi-expert train}}& \multicolumn{2}{c|}{Dice (\%) $\uparrow$} & \multicolumn{2}{c|}{ASSD $\downarrow$} & \multicolumn{2}{c}{95HD $\downarrow$}  \\
            & w/ & w/o            &  w/ & w/o                  &  w/ & w/o \\
            \midrule
            5   & \underline{88.66} & 85.46 &         \underline{1.64} & 2.28  & \underline{5.31} & 8.88 \\
            10  & \underline{88.99} & 87.76 &         \underline{1.59} & 1.83  & \underline{5.12} & 6.39 \\
            15  & \underline{89.18} & 88.47 &         \underline{1.57} & 1.69  & \underline{5.00} & 5.57 \\
            20  & \underline{89.30} & 88.83 & \textbf{\underline{1.55}}& 1.63  & \underline{4.89} & 5.22 \\
            25     & \textbf{89.37} & 89.22 &            \textbf{1.54} & 1.58  &    \textbf{4.81} & 5.01 \\
            30     & \textbf{89.46} & 89.24 &            \textbf{1.53} & 1.56  &    \textbf{4.77} & 4.96 \\
            34     & \textbf{89.48} & 89.37 &            \textbf{1.52} & 1.56  &    \textbf{4.71} & 4.87\\
            \bottomrule
        \end{tabular}
    \end{subtable}
    \vspace{1em}\\
    \begin{subtable}{\linewidth}
        \centering
        \caption{Adapt to $Exp_7$.}
        \begin{tabular}{c|cc|cc|cc}
            \toprule
            \multirow{2}{*}{\diagbox[width=3cm]{\# Samples}{Multi-expert train}}& \multicolumn{2}{c|}{Dice (\%) $\uparrow$} & \multicolumn{2}{c|}{ASSD $\downarrow$} & \multicolumn{2}{c}{95HD $\downarrow$}  \\
                & w/ & w/o            &  w/ & w/o         &  w/ & w/o \\
            \midrule
            5   & \underline{88.41} & 85.30 & \underline{1.58} & 2.09 & \underline{5.45} & 7.78 \\
            10  & \underline{89.30} & 88.16 & \underline{1.43} & 1.65 & \underline{4.89} & 5.92 \\
            15  & \underline{89.78} & 89.28 & \underline{1.36} & 1.45 & \underline{4.56} & 4.95 \\
            20  & \underline{90.21} & 89.92 & \underline{1.30} & 1.37 &             4.36 & 4.47 \\
            25  &             90.41 & 90.33 &             1.27 & 1.30 &             4.25 & 4.36 \\
            30  &    \textbf{90.63} & 90.63 &    \textbf{1.24} & 1.26 &    \textbf{4.14} & 4.13 \\
            34  &    \textbf{90.72} & 90.91 &    \textbf{1.23} & 1.21 &    \textbf{4.11} & 3.91 \\
            \bottomrule
        \end{tabular}
    \end{subtable}
\end{table}

The results are reported in \autoref{tab:n_ann}. It can be observed that, for both new experts, when less than 20 annotated samples are provided, the performance of the expert-adaptive method is significantly better than the randomly initialised counterpart without training with multi-expert annotations, except for the 95HD obtained when adapting the model to $Exp_7$ with 20 samples. Notably, for $Exp_6$, fine-tuning a multi-expert trained model with 20 samples yields comparable performance in terms of mean values (Dice 89.30\%, ASSD 1.55, HD95 4.89) to the best baseline performance obtained using the whole dataset (Dice 89.37\%, ASSD 1.56, HD95 4.87). These observations indicate that the expert-adaptive method can effectively improve the model adaptivity to enable good performance. By contrast, with more than 20 samples for fine-tuning, training with multi-expert annotations tends to yield marginal gain, suggesting that the effectiveness of the expert-adaptive method is more significant under the condition of lightweight fine-tuning where a relatively small number of annotated samples from the new expert is available.

Besides, for both $Exp_6$ and $Exp_7$, when using the method, the significantly best results are obtained with a relatively large number (more than 25 for $Exp_6$ and more than 30 for $Exp_7$) of annotations. This indicates that the most satisfactory performance still cannot be obtained through lightweight fine-tuning with the current method, and there is still room for further improvements to yield more satisfactory performance with fewer samples annotated by the new expert.


\subsection{Influence of the Number of Annotation Experts}
\label{sec:num_experts}

We investigate the influence of the number of experts involved in multi-expert annotation on the model adaptivity by experimenting with the datasets annotated by different numbers of experts. In each experiment, we use a multi-expert dataset annotated by 1, 2, 3, 4, and 5 experts from $Exp_1$-$Exp_5$ to train a segmentation model, respectively, and then fine-tune the model with 10 samples annotated by the new expert $Exp_6$ or $Exp_7$.

\begin{table}[htbp]
    \centering
    \caption{Comparison of segmentation performance when adapting a model trained by the annotated data of different numbers of experts (from $Exp_1$-$Exp_5$) to a new expert ($Exp_6$ or $Exp_7$) with 10 annotated samples. "0" means the model is trained from random initialisation using only the annotations from the new expert. Every experiment yields 10 statistical runs, and the mean metrics of Dice score, ASSD, and 95HD tested on the new expert's annotations are reported. For every metric, the best performer in terms of mean value and those not significantly different from the best (paired t-test with significance level of $\alpha=0.05$), are highlighted in \textbf{bold}.}
    \label{tab:n_exp}
    \begin{subtable}{\linewidth}
        \centering
        \caption{Adapt to $Exp_6$.}
        \begin{tabular}{c|c|c|c}
            \toprule
            \# Experts & Dice (\%) $\uparrow$ & ASSD $\downarrow$ & 95HD $\downarrow$   \\
            \midrule
            0          &         87.76  &         1.83  &         6.39  \\
            \midrule
            1          &         88.66  &         1.65  &         5.38  \\
            2          &         88.89  &         1.60  & \textbf{5.16} \\
            3          & \textbf{88.99} & \textbf{1.59} & \textbf{5.12} \\
            4          & \textbf{89.00} & \textbf{1.58} & \textbf{4.97} \\
            5          & \textbf{89.13} & \textbf{1.58} & \textbf{4.95} \\
            \bottomrule
        \end{tabular}
    \end{subtable}
    \vspace{1em}\\
    \begin{subtable}{\linewidth}
        \centering
        \caption{Adapt to $Exp_7$.}
        \begin{tabular}{c|c|c|c}
            \toprule
            \# Experts & Dice (\%) $\uparrow$& ASSD $\downarrow$& 95HD $\downarrow$ \\
            \midrule
            0          &         88.16  &         1.65  &         5.92  \\
            \midrule
            1          &         89.11  &         1.47  &         5.03  \\
            2          & \textbf{89.24} & \textbf{1.45} & \textbf{4.89} \\
            3          & \textbf{89.30} & \textbf{1.43} & \textbf{4.89} \\
            4          & \textbf{89.38} & \textbf{1.41} & \textbf{4.77} \\
            5          & \textbf{89.50} & \textbf{1.39} & \textbf{4.71} \\
            \bottomrule
        \end{tabular}
    \end{subtable}
\end{table}

The results are shown in \autoref{tab:n_exp}. It can be observed that involving the annotations of more than 3 experts yields consistently better adaptation performance than the single-expert counterpart, confirming that multi-expert annotation can improve the model adaptivity to a new expert with lightweight fine-tuning. Besides, the performance monotonically improves with a larger number of experts annotating the dataset in the training stage, with 5-expert annotation yielding the top performance in terms of mean values. This makes sense because more experts provide more diversified segmentation annotations for the model to learn, thus further reducing its bias towards individual experts and improving its adaptivity.
As the number of experts increases, the t-test results show that the significance of the performance gains cap early: for $Exp_6$, involving 3 or more experts in the training stage does not yield significantly better performance, and for $Exp_7$, involving 2 or more experts yields no significant difference. However, the mean performance values often improve steadily. This may suggest a potential room for further improvements by more effectively utilising the annotations from more experts and using more diversified annotations.

\section{Conclusions and Future Work}
\label{conclusions}
We evaluated an existing expert-adaptive MIS method with customisation made to accommodate the limited data availability and mobility in practice, which is composed of multi-expert annotation, multi-task DNN-based model training, and lightweight model fine-tuning on the limited annotated data from a new expert to address the inherent variation in expert-specific annotations for enhancing the adaptivity of the MIS model to a new expert. Experiments on the brain MRI segmentation tasks with limited training data demonstrate its effectiveness and the impact of its key parameters. In the future, we plan to comprehensively compare the original and customised versions of the expert-adaptive method \cite{nichyporuk2022rethinking} in different practical scenarios, investigate how to make best use of the annotations from multiple experts during model training by using different training methods, e.g., by leveraging evolutionary multi-task optimisation based approaches \cite{song2019multitasking, wu2022evolutionary}. Also, we will study the contributions of the annotations from different experts to the adaptivity of the trained model, e.g., via graph matching \cite{gong2016discrete} and learning vector quantisation \cite{qin2005initialization} techniques, aiming to find more diversified experts to create more effective multi-expert annotation.

\section*{Acknowledgment}
This work was performed on the OzSTAR national facility at Swinburne University of Technology. The OzSTAR program receives funding in part from the Astronomy National Collaborative Research Infrastructure Strategy (NCRIS) allocation provided by the Australian Government, and from the Victorian Higher Education State Investment Fund (VHESIF) provided by the Victorian Government. This work was supported in part by the Peter MacCallum Cancer Centre and the Australian Research Council (ARC) under LP180100114 and DP200102611.

\bibliographystyle{ieeetr}
\bibliography{refers}

\end{document}